\documentclass[11pt]{article}

\usepackage[final]{acl}

\usepackage{times}
\usepackage{latexsym}
\usepackage{placeins}
\usepackage[T1]{fontenc}
\usepackage{booktabs}
\usepackage{tabularx}
\usepackage{caption}
\usepackage{float}
\usepackage{longtable}
\usepackage[most]{tcolorbox} % 'most' usually includes breakable
\tcbuselibrary{breakable}    % Explicitly load the library

\usepackage[utf8]{inputenc}

\usepackage{tikz}
\usepackage{xcolor}

\definecolor{mygreen}{HTML}{2E7D32}      % A nice, dark green for text
\definecolor{mygreenbg}{HTML}{E8F5E9}    % A very light green for background
\definecolor{myorange}{HTML}{E87400}    % A strong orange for text
\definecolor{myorangebg}{HTML}{FEF3DE}  % A light yellow/orange for background
\definecolor{mygray}{HTML}{546E7A}      % A neutral, dark gray for text
\definecolor{mygraybg}{HTML}{CFD8DC}    % A 

\newcommand{\diffbox}[4]{%
    \small % Use a slightly smaller font inside the box
    \tikz[baseline=(char.base)]\node[
        rounded corners=3pt,    % Rounded edges
        fill=#1,                % Background color
        text=#2,                % Text color
        inner sep=2pt,          % Padding inside the box
        font=\bfseries,         % Bold text
        ] (char) {#3 #4};%
}

\newcommand{\goodup}[1]{\diffbox{mygreenbg}{mygreen}{$\uparrow$}{#1}}

\usepackage{microtype}
\usepackage{inconsolata}
\usepackage{amsmath}
\usepackage{graphicx}

\title{Context Volume Drives Performance: Tackling Domain Shift in Extremely Low-Resource Translation via RAG}

\author{
  \textbf{David Samuel Setiawan}, 
  \textbf{Raphaël Merx}, 
  \textbf{Jey Han Lau} \\
  The University of Melbourne \\
  \small{
    \texttt{\{david.setiawan, raphael.merx\}@student.unimelb.edu.au}
  } \\
  \small{
    \texttt{laujh@unimelb.edu.au}
  }
}

\begin{document}
\maketitle
\begin{abstract}
Neural Machine Translation (NMT) models for low-resource languages suffer significant performance degradation under domain shift. We quantify this challenge using \textbf{Dhao}, an indigenous language of Eastern Indonesia with no digital footprint beyond the New Testament (NT). When applied to the unseen Old Testament (OT), which exhibits a 3x increase in OOV rate (25.9\%) and distinct thematic divergence from the New Testament (NT) training data, a standard NMT model fine-tuned on the NT drops from an \textbf{in-domain score of 36.17 chrF++} to \textbf{27.11 chrF++}. To recover this loss, we introduce a \textbf{hybrid framework} where a fine-tuned NMT model generates an initial draft, which is then refined by a Large Language Model (LLM) using Retrieval-Augmented Generation (RAG). The final system achieves \textbf{35.21 chrF++} (+8.10 recovery), effectively matching the original in-domain quality. Our analysis reveals that this performance is driven primarily by the \textbf{number of retrieved examples} rather than the choice of retrieval algorithm. Qualitative analysis confirms the LLM acts as a robust ``safety net,'' repairing severe failures in zero-shot domains.
\end{abstract}

\section{Introduction} \label{S1}

% \subsection{The Hook}
For the majority of the world's 7,000+ languages, biblical texts often represent the only available large-scale digital resource \citep{ranathunga2023neural}. However, translation efforts typically prioritize the New Testament (NT), leaving the Old Testament (OT), which constitutes 75\% of the Bible, untranslated. The reason is that the NT contains the theological core and is an essential starting point for a new believer or a new church. Furthermore, the linguistic complexity and size of the OT present significant translation challenges.

As of August 2025, while 2,574 languages possess a complete NT, only 776 have a complete OT \citep{Wycliffe2025Stats}. While leveraging available NT data to train Machine Translation (MT) models for the OT is a logical next step, this workflow presents a distinct \textbf{domain shift} challenge. Despite forming a single canon, the NT and OT diverge significantly in vocabulary and style, as they originate from different source languages (Hebrew vs. Koine Greek) and cover distinct themes (historical narrative vs. theological discourse).

Our analysis of the English source text (World English Bible) quantifies this shift: the Out-of-Vocabulary (OOV) rate relative to the NT training vocabulary increases from 8.1\% on the in-domain NT validation set to 25.9\% on the OT test set (see Appendix \ref{sec:domain_shift_analysis}). Consequently, NMT models trained solely on the NT generalize poorly, leading to marked performance degradation \citep{akerman2023ebible}.

In this work, we address this NT-to-OT shift using \textbf{Dhao}, an indigenous language of Eastern Indonesia with fewer than 5,000 speakers  \citep{Ethnologue2025}. Unlike existing domain adaptation work which relies on target-domain monolingual corpora \cite{chu2020survey, marashian-etal-2025-priest}, we operate under a stricter constraint: the primary training data is domain-bound (NT), with the only available general-domain signal coming from a small digitized grammar book. To address this, we introduce a hybrid \textbf{NMT + LLM Post-Editing} framework. We utilize a fine-tuned NMT model to generate an initial draft and employ a Retrieval-Augmented Generation (RAG) enhanced LLM to refine the output using context retrieved from both the grammar book and the NT.

We systematically compare \textbf{sentence-level retrieval} (whole-sentence similarity) against \textbf{word-level retrieval} (aggregated source word matches) to assess their impact on the proposed hybrid pipeline. Our results indicate that while increasing context volume consistently yields gains across all strategies, the specific choice of retrieval algorithm is secondary. The final optimized system restores character-level overlap (chrF++) to in-domain levels, though a gap remains in subword-level similarity (spBLEU), likely due to the higher stylistic and lexical divergence of the Old Testament. Based on these findings, we summarize our primary contributions as follows:

\paragraph{Contributions}
\begin{enumerate}
    \item \textbf{Zero-Shot Domain Adaptation for Unseen Languages:} We propose a hybrid NMT+LLM framework that successfully tackles domain shift for an extremely low-resource language with no digital footprint. We demonstrate that this architecture allows an LLM to correct a language it has never seen (Dhao) by leveraging the structural priors of a fine-tuned NMT model, outperforming baselines that rely on either model individually.
    \item \textbf{Context Volume as the Primary Performance Driver:} We demonstrate that context volume drives LLM post-editing performance more significantly than the choice of retrieval algorithm in low-resource RAG. Our analysis shows that distinct retrieval strategies yield comparable gains when normalized for volume. 
    \item \textbf{Validation of the ``Safety Net`` Hypothesis:} We provide qualitative evidence that LLM post-editing specifically mitigates catastrophic NMT failures, such as hallucinations and repetition loops, in zero-shot domains, validating the hybrid architecture's robustness for data-scarce settings.
\end{enumerate}

\section{Related Work}

\paragraph{Domain Shift in Low-Resource MT}
Standard NMT models are highly lexicalized, making them brittle when applied to distributions differing from their training data \citep{koehn-knowles-2017-six,hu-etal-2019-domain-adaptation,haddow-etal-2022-survey}. This brittleness often manifests as catastrophic hallucinations or fluent but unfaithful outputs when the model encounters out-of-domain data \cite{muller-etal-2020-domain, raunak-etal-2021-curious}. Specifically, such shifts have been shown to exacerbate the model's reliance on training-domain priors, leading it to ignore source constraints in favor of frequently observed sequences from the training set \cite{wang-sennrich-2020-exposure}. 

\paragraph{LLMs and the Hybrid Solution}
While Large Language Models (LLMs) excel at high-resource translation, they struggle with ``unseen'' languages due to a lack of pre-training exposure \citep{robinson-etal-2023-chatgpt,hendy_how_2023}. Effective translation often remains unattainable for languages with underrepresented scripts even with RAG \citep{lin-etal-2025-large}. However, \textbf{In-Context Learning (ICL) with language alignment} (e.g., dictionary constraints) has been identified as a viable method to unlock LLM capabilities for these languages, significantly outperforming fine-tuning which suffers from overfitting \citep{li-etal-2025-context-learning}.

To mitigate the weaknesses of both paradigms, recent literature converges on hybrid architectures. An NMT model followed by an LLM post-editor has been shown to be an optimal recipe for low-resource translation, specifically for mitigating ``lexical confusion'' \citep{nielsen-etal-2025-alligators}. However, the optimal retrieval granularity for this post-editing remains an open question. While current strategies optimize for n-gram diversity \citep{caswell-etal-2025-smol} or compositional phrases \citep{zebaze-etal-2025-compositional}, word-level retrieval has also been proposed for grammatical learning \citep{tanzer_benchmark_2024}. Our work synthesizes these insights: we employ the hybrid framework validated by \citet{nielsen-etal-2025-alligators}, but we systematically compare these sentence-level versus word-level strategies to determine whether performance gains stem from choice of retrieval algorithm or simply the increased volume of in-context examples.

\section{Experimental Setup}

\subsection{Data Construction}
We utilize the Dhao language resources introduced in Section \ref{S1} to construct a zero-shot domain adaptation benchmark. As Dhao lacks a standard digital footprint, we curate our datasets from the only two available sources: a Bible translation and a digitized grammar book \citep{balukh2016grammar}.

\paragraph{Primary Corpus (Parallel Bible)}
We source the parallel biblical text from the \textbf{ebible corpus} \citep{akerman2023ebible}. We align the \textbf{target} Dhao translation (written in \textbf{Latin script}) against the \textbf{source} World English Bible (WEB). The primary objective of this alignment is to decompose the raw verse-level text into a sentence-level parallel corpus, thereby providing the granular signal required for effective NMT training. The data is partitioned as follows:
\begin{itemize}
    \item \textbf{In-domain (train \& eval):} The complete New Testament (NT), comprising 7,644 parallel verses. We reserve 95\% for fine-tuning the NMT model and 5\% for in-domain validation.
    \item \textbf{Out-of-domain (test):} The first 500 verses of the Book of Genesis (Old Testament). Although the Old Testament is not fully translated in Dhao, a translation of Genesis exists; we utilize this text as our \textbf{ground truth} for evaluation. It serves as a strictly \textbf{unseen domain} to evaluate the model's generalization capabilities under the lexical shift described in Section \ref{S1}. We verified that none of these verses appear in the supplementary grammar book, which exclusively cites examples from the New Testament \citep{balukh2016grammar}, ensuring no data leakage occurs.
\end{itemize}

\paragraph{Supplementary Corpus (Grammar Extraction)}
To support RAG-based post-editing, we extracted a supplementary corpus from \textit{A Grammar of Dhao} \citep{balukh2016grammar}. Using a semi-automated pipeline involving PDF segmentation and LLM extraction (detailed in Appendix \ref{sec:data_construction}), we curated a clean dataset of \textbf{1,011 parallel sentences} and \textbf{2,377 bilingual lexicon entries}. These resources represent the only available general-domain data for the language.

\subsection{Models}
We employ a hybrid architecture that leverages the complementary strengths of specialized NMT and general-purpose LLMs:
\begin{itemize}

    \item \textbf{NMT (Drafting):} We use \textbf{NLLB-200-distilled-600M} \citep{costa2022no}. We specifically select this distilled version over larger variants (e.g., 1.3B or 3.3B) to prioritize faster inference speeds. This allows for rapid iteration during experimentation while still leveraging the model's massive multilingual pre-training, which provides a robust initialization for fine-tuning on the limited Dhao NT data.
    \item \textbf{LLM (Post-Editing):} We utilize \textbf{Gemini 2.5 Flash} \citep{comanici2025gemini} for the post-editing stage. This model was selected for its large context window (enabling the ingestion of extensive RAG examples) and its cost-effectiveness for iterative experimentation.
\end{itemize}

\subsection{Evaluation Metrics}
Given the low-resource nature of Dhao and the lack of standardized linguistic tools (e.g., morphological analyzers or tokenizers), we report performance using two robust metrics:
\begin{itemize}
    \item \textbf{spBLEU:} A tokenizer-agnostic BLEU score using SentencePiece \citep{costa2022no}. Since Dhao lacks a gold-standard tokenizer, spBLEU ensures that performance is measured based on learned sub-word units rather than potentially flawed rule-based tokenization.
    \item \textbf{chrF++:} A character n-gram metric \citep{popovic-2017-chrf}. We prioritize chrF++ as it is strictly more robust than word-level BLEU for low-resource languages. By measuring character-level overlap, it provides partial credit for correct stems even when the model generates incorrect affixes or spelling variations, which is critical for accurately evaluating an unseen dialect like the Old Testament.
\end{itemize}

\section{Methodology}
We propose a hybrid translation framework that integrates specialized NMT with LLM-based post-editing to mitigate domain shift in extremely low-resource settings. While the architecture follows a standard post-editing setup, our primary contribution lies in the systematic optimization of the RAG context, when translating an unseen language with domain shift.

\subsection{The Hybrid NMT-LLM Pipeline}
\begin{figure*}[t]
    \centering
    % Ensure filename matches your upload
    \includegraphics[width=1.0\textwidth]{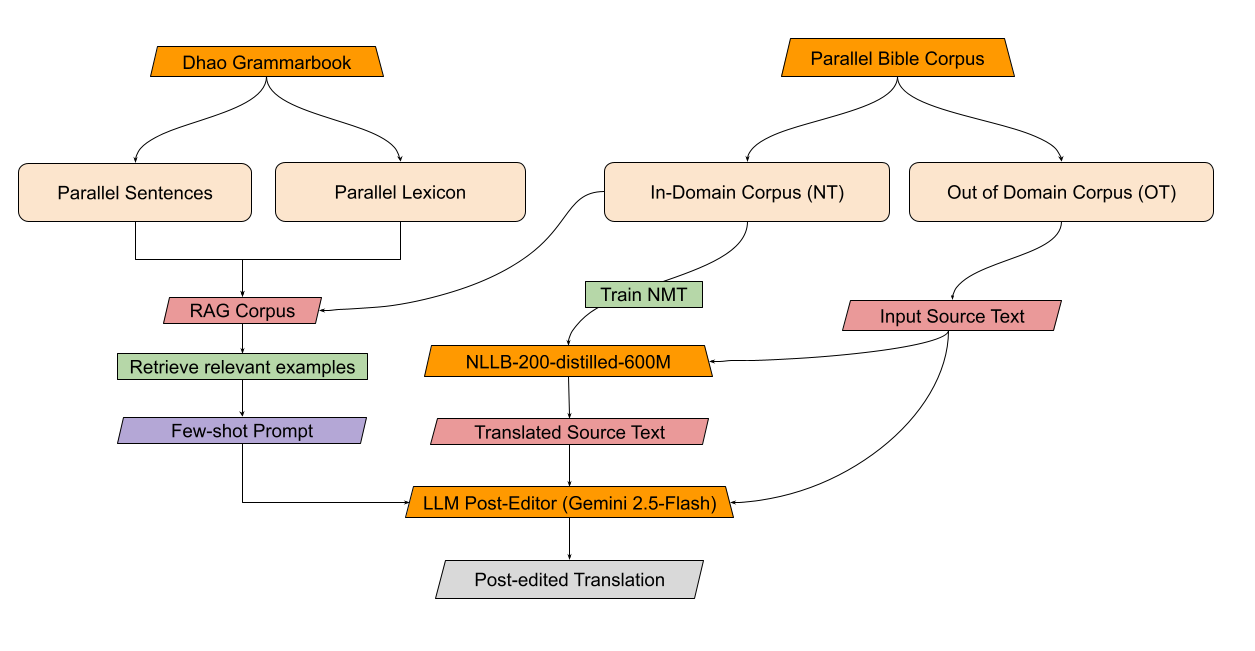}
    \caption{\textbf{Hybrid Post-Editing Architecture.} The workflow integrates two parallel streams. \textbf{Center:} The NLLB model, fine-tuned on the In-Domain (NT) corpus, generates an initial \textit{Translated Source Text} (draft). \textbf{Left:} A Retrieval-Augmented Generation (RAG) module queries the combined corpus (Grammar Book + NT) to extract relevant examples for the \textit{Few-shot Prompt}. \textbf{Bottom:} The LLM Post-Editor (Gemini) synthesizes the NMT draft, the original source, and the retrieved context to produce the final \textit{Post-edited Translation}. \textbf{Legend:} Orange: Core objects (data and models) $|$ Green: Processing steps $|$ Red: Intermediate outputs $|$ Purple: Prompt configuration $|$ Gray: Final output.}
    \label{fig:post_editing_pipeline}
\end{figure*}

The translation pipeline, illustrated in Figure \ref{fig:post_editing_pipeline}, consists of two distinct phases:

\paragraph{Phase 1: NMT Drafting}
We fine-tune the NMT model described in Section 3.2 on the in-domain (NT) corpus to generate an initial hypothesis $y_{nmt}$. We adopt the optimal hyperparameters from the eBible benchmark \citep{akerman2023ebible}, as detailed in Appendix \ref{sec:appendix_hyperparams}.

\paragraph{Phase 2: RAG-Enhanced Post-Editing}
We post-edit the translation with \textbf{Gemini 2.5 Flash} \citep{comanici2025gemini}. The LLM receives a structured prompt containing: (1) the original source sentence $x$; (2) the NMT draft $y_{nmt}$; (3) a set of retrieved parallel sentences (sourced from both the NT and the grammar book); and (4) a set of retrieved lexicon entries formatted as direct mappings (e.g., \textit{English Word (POS) $\rightarrow$ Dhao Word}). The composition of these retrieved contexts depends on the experimental setup. We evaluate parallel sentence retrieval and lexicon retrieval independently, but combine them in the final optimized system (see Section \ref{sec:final_results}). The full prompt structure and integration details are provided in Appendix \ref{sec:prompts}. The model is instructed to compare the draft against the source and selectively correct NMT failures like hallucinations or repetition loops only when necessary rather than re-translating from scratch.

% \paragraph{Phase 2: RAG-Enhanced Post-Editing}
% We post-edit the translation with \textbf{Gemini 2.5 Flash} \citep{comanici2025gemini}. The LLM receives a structured prompt containing:
% \begin{enumerate}
%     \item The original source sentence $x$.
%     \item The NMT draft $y_{nmt}$.
%     \item A set of retrieved parallel sentences (sourced from both the NT and the grammar book).
%     \item A set of retrieved lexicon entries formatted as direct mappings (e.g., \textit{English Word (POS) $\rightarrow$ Dhao Word}).
% \end{enumerate}
% The composition of these retrieved contexts depends on the experimental setup. We evaluate parallel sentence retrieval and lexicon retrieval independently, but combine them in the final optimized system (see Section \ref{sec:final_results}). The full prompt structure and integration details are provided in Appendix \ref{sec:prompts}. The model is instructed to compare the draft against the source and correct specific errors (e.g., hallucinations, repetition loops) only when necessary, rather than re-translating from scratch.

\subsection{Baselines}
To evaluate the proposed framework, we compare against four baseline configurations evaluated on the out-of-domain (OT) test set. These baselines rely on \textbf{static retrieval} strategies, contrasting with the dynamic retrieval methods detailed in Section \ref{sec:retrieval_strategies}.

\begin{enumerate}
    \item \textbf{NMT-Only:} The NLLB model fine-tuned solely on the NT data, serving as the lower-bound for domain adaptation.
    \item \textbf{NMT + Grammar:} The NLLB model fine-tuned on the NT data augmented with the grammar book parallel sentences.
    \item \textbf{LLM Direct Translation:} Gemini 2.5 translating directly from English to Dhao in 0-shot (no context) and 5-shot (5 fixed, randomly selected NT sentences) settings
    \item \textbf{LLM Post-Editing (No RAG):} The hybrid pipeline without retrieval, relying on internal LLM knowledge to correct the NMT draft in 0-shot and 5-shot settings.
\end{enumerate}

\subsection{Retrieval Strategies (RAG)}
\label{sec:retrieval_strategies}
A core contribution of this work is investigating whether performance gains in low-resource RAG stem from retrieval strategy, on the volume of examples, or both. Unlike the baselines which use static context, these strategies dynamically retrieve examples relevant to the specific input sentence $x$.

\subsubsection{Parallel Sentence Retrieval}
We retrieve relevant parallel pairs from a combined corpus of the in-domain NT and the grammar book. All retrieval operations are performed on the source side (English), bypassing the need for retrieval models trained on Dhao. We evaluate four strategies:

\paragraph{Sentence-Level Approaches (Fixed $k$)}
These methods retrieve a fixed number of sentences $k$ based on their similarity to the source input. We test $k \in \{5, ..., 100\}$.
\begin{itemize}
    \item \textbf{BM25 (lexical):} A standard sparse retrieval method that ranks sentences based on exact keyword overlap, normalized for document length.
    \item \textbf{BGE Embeddings (semantic):} A semantic retrieval method using \texttt{bge-large-en-v1.5} \cite{bge_embedding}. We compute the cosine similarity between the source sentence embedding and corpus embeddings to capture semantic relevance beyond keyword matching. 
    \item \textbf{ChrF-Counterweighted (lexical, with diversity focus):} Adapted from \citet{caswell-etal-2025-smol}, this method promotes n-gram diversity. It iteratively selects examples with high character n-gram overlap while penalizing n-grams present in previously selected examples, ensuring the context window is not filled with redundant phrasing.
\end{itemize}

\paragraph{Word-Level Approach (Dynamic $k$)}
Inspired by \citet{tanzer_benchmark_2024}, we implement a \textbf{Fuzzy Word Matching} strategy. Instead of retrieving based on the whole sentence, we retrieve the top-$n$ parallel sentences for each word in the source sentence. We compute token similarity using the normalized Levenshtein distance via the \texttt{rapidfuzz} library, retaining only matches with similarity $\ge 0.5$. Unlike sentence-level methods, the total number of examples $k$ is \textbf{dynamic}, scaling with the sentence length ($k \approx n \times \text{sentence\_length}$). We ablate $n \in \{1, 2, 3, 5, 10, 15, 20\}$ to determine if granular, word-level context outperforms sentence-level retrieval.

\subsubsection{Lexicon Retrieval}
We further augment the context with bilingual dictionary entries extracted from the grammar book. We compare two configurations:
\begin{itemize}
    \item \textbf{Fuzzy Retrieval:} Retrieving the top-$n$ similar lexicon entries per source word. We evaluate $n \in \{3, 5, 10, 15, 20, 25, 30, 50, 70, 100\}$.
    \item \textbf{Full Dictionary:} Providing the entire 2,375-entry lexicon in the context window, treating the dictionary as a static resource rather than a retrieved element.
\end{itemize}

\section{Results}

\subsection{Baseline Performance \& Domain Shift}
\label{sec:baseline_results}
We first quantify the severity of the domain shift by evaluating the fine-tuned NMT model on the out-of-domain (OT) test set. As shown in Table~\ref{tab:baseline_performance}, the model suffers a performance collapse: spBLEU drops from \textbf{25.19} (in-domain NT) to \textbf{7.66} (OT), and chrF++ drops from \textbf{36.17} to \textbf{27.11}. This confirms that standard fine-tuning is insufficient for the lexical and stylistic divergence of the NT-to-OT shift.

% TABLE: BASELINE PERFORMANCE
\begin{table}[t]
    \centering
    \resizebox{\columnwidth}{!}{%
    \begin{tabular}{llcc}
        \toprule
        \textbf{Model} & \textbf{Context} & \textbf{spBLEU} & \textbf{chrF++} \\
        \midrule
        \multicolumn{4}{l}{\textit{In-Domain Reference (NT)}} \\
        NMT (NLLB) & None & 25.19 & 36.17 \\
        \midrule
        \multicolumn{4}{l}{\textit{Out-of-Domain Test (OT)}} \\
        NMT (NLLB) & None & 7.66 & 27.11 \\
        NMT + Grammar & None & 7.67 & 26.62 \\
        LLM Direct & 0-shot & 2.98 & 18.84 \\
        LLM Direct & 5-shot & 7.37 & 22.95 \\
        LLM Post-Edit & 0-shot & 10.65 & 27.94 \\
        \textbf{LLM Post-Edit} & \textbf{5-shot} & \textbf{12.54} & \textbf{29.62} \\
        \bottomrule
    \end{tabular}%
    }
    \caption{\textbf{Baseline Performance Quantification.} Comparison of NMT and LLM baselines on the out-of-domain (OT) test set. The first row shows in-domain (NT) performance as a ceiling reference. Note the severe drop when NMT is applied to the OT.}
    \label{tab:baseline_performance}
\end{table}

\paragraph{LLMs as a Safety Net}
Gemini-2.5-Flash failed as a direct translator (2.98 spBLEU), confirming it possesses no prior knowledge of Dhao. However, the \textbf{Hybrid Post-Editing} framework significantly outperformed the NMT baseline (+4.88 spBLEU in the 5-shot setting). Qualitative analysis reveals that the LLM acts as a ``safety net.'' The NMT model frequently suffers from catastrophic failures on OOV terms, such as entering infinite repetition loops. The LLM consistently identifies and truncates these loops, recovering coherent text (see Table~\ref{tab:qualitative_examples}).

\subsection{Retrieval Strategy Analysis}
\label{sec:retrieval_analysis}
A core research question was whether performance gains in low-resource RAG stem from the choiceß of retrieval algorithm (e.g., semantic embeddings vs. lexical overlap) or simply the volume of in-context examples. To answer this, we decouple our analysis into two parts: first evaluating parallel sentence retrieval in isolation, and subsequently analyzing the impact of lexicon retrieval.

\subsubsection{Parallel Sentence Retrieval Analysis}
\label{sec:sent_retrieval_analysis}

% FIGURE: RETRIEVAL CURVES (Sentence vs Word Level)
\begin{figure*}[t]
    \centering
    % Ensure filename matches your upload: image_81afc7.png
    \includegraphics[width=\textwidth]{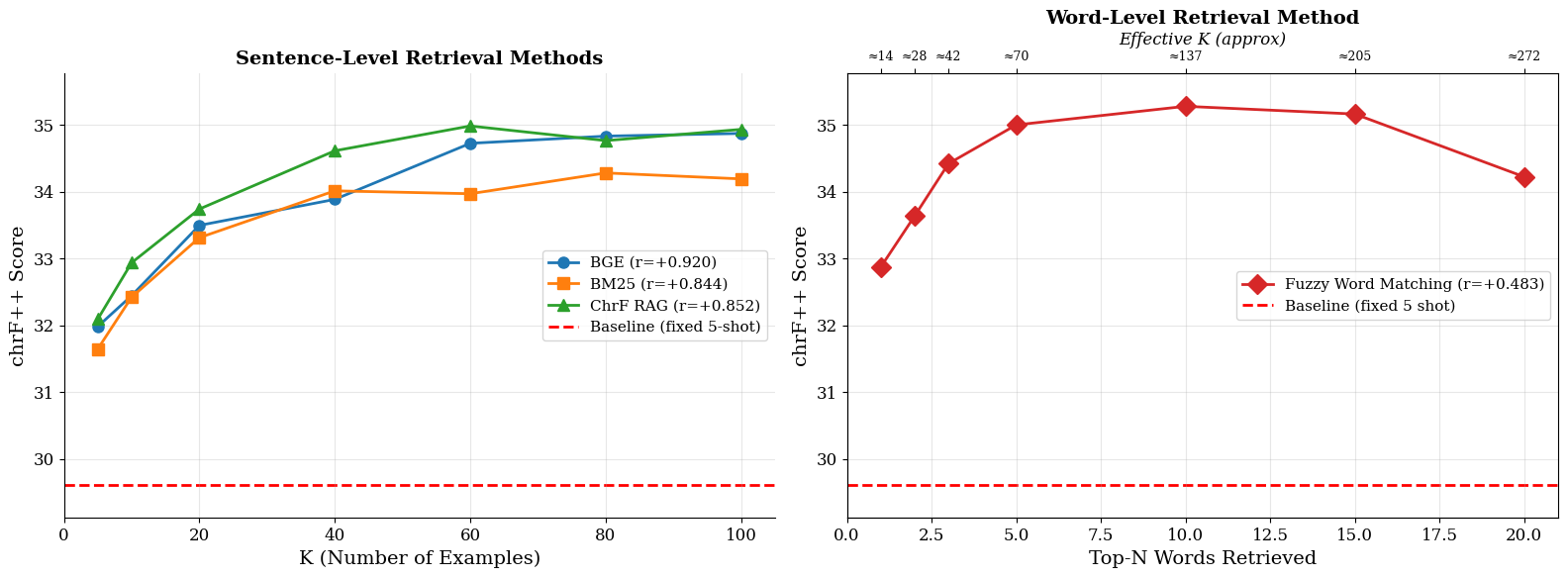}
    \caption{\textbf{Impact of Context Volume on Performance.} Comparison of absolute chrF++ scores across retrieval strategies relative to the \textbf{Fixed 5-Shot Baseline} (dashed red line, corresponding to the LLM Post-Editing baseline in Table~\ref{tab:baseline_performance}). \textbf{Left:} Sentence-level methods (BGE, BM25, ChrF-RAG) show rapid initial gains but plateau at $K \approx 60$. \textbf{Right:} The Word-Level strategy allows the model to ingest a higher effective volume of examples (peaking at effective $K \approx 137$) to squeeze out marginal performance gains.}
    \label{fig:retrieval_curves}
\end{figure*}

\paragraph{Impact of Context Volume}
As shown in Figure \ref{fig:retrieval_curves} (Left) and detailed in Table \ref{tab:app_sentence_results} (Appendix \ref{sec:appendix_results}), all dynamic sentence-level strategies outperform the static 5-shot baseline (dashed red line) immediately, even at low $k$. We observe strong, consistent improvement as the context volume increases from 5 to 60, regardless of whether dense embeddings (BGE) or sparse matching (BM25) is used. However, these methods plateau around $k \approx 60$. In contrast, the Word-Level strategy (Figure \ref{fig:retrieval_curves}, Right) circumvents this saturation. By retrieving granular examples, it allows the model to effectively utilize a much larger context volume, with performance continuing to scale until peaking at an effective $k \approx 137$ (see Table \ref{tab:app_word_results} for full numerical results).

\begin{figure}[t]
    \centering
    \includegraphics[width=\columnwidth]{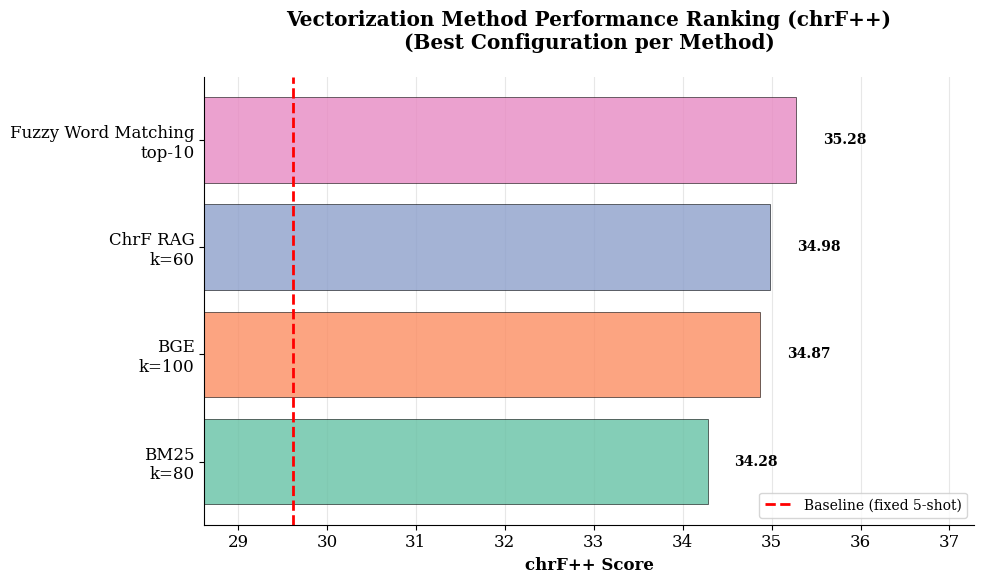}
    \caption{\textbf{Retrieval Strategy Performance Convergence.} We compare the optimal configuration of the Word-Level strategy (Fuzzy Matching, top-10) against the best configurations of three sentence-level baselines: ChrF-RAG ($k=60$), BGE Semantic Retrieval ($k=100$), and BM25 ($k=80$). The bar chart displays the absolute chrF++ scores, with the dashed red line indicating the Fixed 5-shot Baseline.}
    \label{fig:retrieval_ranking}
\end{figure}

\paragraph{Performance Convergence and Efficiency Trade-offs}
When comparing the optimal configurations of each method, we observe a convergence in peak performance. As illustrated in the bar chart in Figure \ref{fig:retrieval_ranking}, the maximum chrF++ scores for the Word-Level, ChrF-RAG, and BGE strategies are all within \textbf{0.5 points} of each other. The \textbf{Word-Level Fuzzy Matching} strategy achieves the absolute highest score (\textbf{35.28 chrF++}), but it outperforms the best sentence-level baseline (ChrF-RAG: 34.98) by only a small margin ($+0.3$).

However, efficiency analysis favors sentence-level retrieval. ChrF-RAG achieves 99\% of the optimal performance with just $K=60$ examples, whereas the word-level strategy requires nearly double the volume ($\approx$137) for a marginal gain. This makes sentence-level retrieval the more pragmatic choice for production environments where token usage and latency are constraints.

% FIGURE: BAR CHART RANKING

\paragraph{Impact of Retrieval Corpus}
To validate the generalizability of our findings, we isolated the impact of the supplementary grammar data. We compared the performance of the best configuration (Word-Level, $n=10$) using the combined corpus versus using \textit{only} the in-domain NT for retrieval.

Results show that restricting the retrieval source to the NT corpus results in only a marginal performance drop compared to the combined corpus (from \textbf{35.28} to \textbf{35.01 chrF++}, and \textbf{18.93} to \textbf{18.47 spBLEU}).
This confirms that the approach remains a viable solution for extremely low-resource languages where a Bible translation may be the \textit{only} available digital resource, without requiring the digitization of supplementary grammar books.

\subsubsection{Lexicon Retrieval Analysis}
\label{sec:lexicon_retrieval_analysis}
Having analyzed parallel sentence retrieval, we independently evaluate the utility of the bilingual lexicon.

\paragraph{Impact of Dictionary Volume}

\begin{figure}[t]
    \centering
    % Ensure filename matches your upload: image_81afa5.png
    \includegraphics[width=\columnwidth]{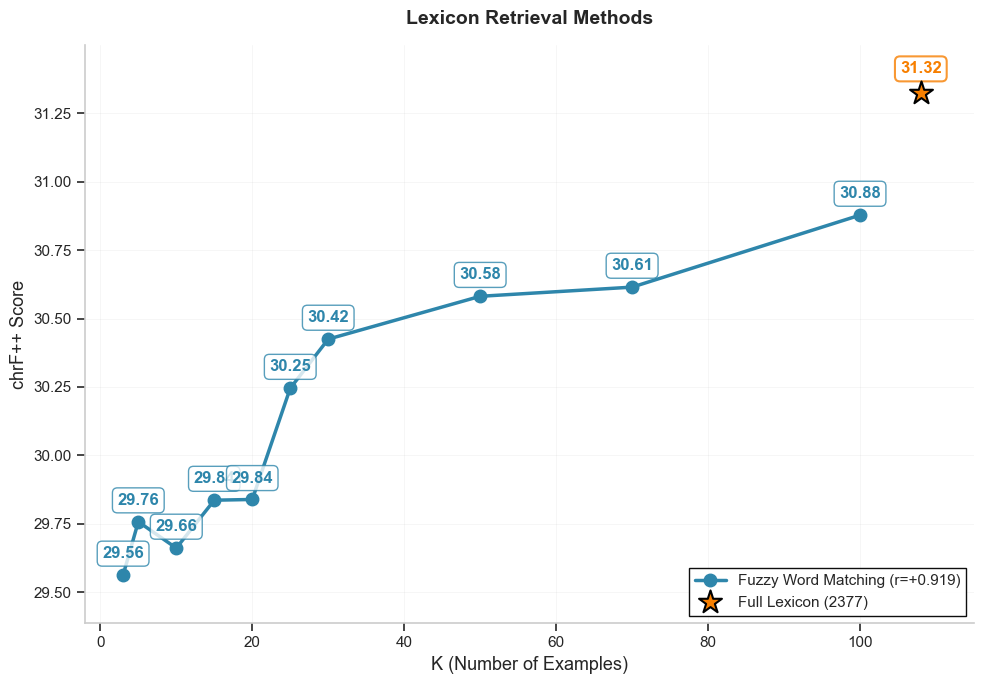}
    \caption{\textbf{Lexicon Retrieval Performance.} Impact of increasing the number of retrieved lexicon entries ($K$) on post-editing performance. Unlike sentence-level retrieval which plateaus, lexicon retrieval improves monotonically with volume. The highest performance is achieved by providing the \textbf{Full Lexicon} (star marker), yielding a chrF++ score of 31.32.}
    \label{fig:lexicon_curve}
\end{figure}

As shown in Figure \ref{fig:lexicon_curve}, performance improves linearly with the number of entries provided, contrasting with the plateau observed in sentence retrieval. The optimal performance was achieved by providing the \textbf{Full Dictionary}, yielding \textbf{16.27 spBLEU} (+8.61) and \textbf{31.32 chrF++} (+4.21). The fuzzy matching approach with $N=100$ closely approximated this peak ($\approx$30.88 chrF++), suggesting that for targeted lexical information, quantity is strictly beneficial. Providing the entire lexicon maximizes the probability of retrieving precise translations for OOV terms without introducing the syntactic noise inherent in full sentences (see Table \ref{tab:app_lexicon_results} in Appendix \ref{sec:appendix_results} for full results).

% FIGURE: LEXICON CURVE

\subsection{Final System Performance}
\label{sec:final_results}
Our final system combines the optimal parallel sentence retrieval strategy, the Word-Level Fuzzy Matching ($n=10$, yielding an effective $k \approx 137$), with the full bilingual lexicon. As shown in Table \ref{tab:final_results}, this yields the highest overall translation accuracy.

% TABLE: FINAL RESULTS (Refined Caption)
\begin{table}[t]
    \centering
    \resizebox{\columnwidth}{!}{%
    \begin{tabular}{l l l}
        \toprule
        \textbf{Configuration} & \textbf{spBLEU} & \textbf{chrF++} \\
        \midrule
        \textit{In-Domain Reference (NT)} & \textit{25.19} & \textit{36.17} \\
        \midrule
        \multicolumn{3}{l}{\textit{Out-of-Domain Results (Genesis)}} \\
        Baseline (NMT Only) & \phantom{1}7.66 & 27.11 \\
        + Lexicon (Full) & 16.27 \goodup{8.61} & 31.32 \goodup{4.21} \\
        + Sentences (Word-Level) & 18.93 \goodup{11.27} & \textbf{35.28} \goodup{8.17} \\
        \textbf{+ Combined (Final)} & \textbf{19.88} \goodup{12.22} & 35.21 \goodup{8.10} \\
        \bottomrule
    \end{tabular}%
    }
    \caption{\textbf{Experimental Results.} Comparison of component contributions. The first row provides the in-domain (NT) upper bound. Our final combined system (Word-Level Sentences + Full Lexicon) achieves \textbf{35.21 chrF++}, effectively recovering the performance lost to domain shift by nearly matching the in-domain reference of 36.17.}
    \label{tab:final_results}
\end{table}

The final model achieves \textbf{35.21 chrF++}, which almost matches the in-domain performance of the NMT model (36.17 chrF++). This indicates that our RAG-enhanced post-editing framework has successfully recovered the performance lost to domain shift.

Interestingly, while the combined approach yields the highest spBLEU (\textbf{19.88}), the chrF++ score (\textbf{35.21}) is slightly lower than the sentence-only configuration (\textbf{35.28}). This suggests a nuanced trade-off in metric sensitivity: the lexicon provides high-precision constraints that improve exact \textbf{subword-level overlap} (boosting spBLEU), which is otherwise hindered by the 25.9\% OOV rate. Conversely, the \textbf{character-level overlap} (chrF++) appears to reach saturation with sentence-level retrieval alone, nearly matching the in-domain reference of 36.17. The slight drop in chrF++ when adding the full lexicon may indicate that the increased \textbf{context volume} introduces minor syntactic noise that outweighs marginal gains in character-level accuracy. Nevertheless, the combined model remains the most robust overall system for recovering performance across both subword and character granularities.

\subsection{Qualitative Analysis}
\label{sec:qualitative_analysis}

To investigate the source of the performance gains, we analyzed the test set outputs. We found that the fine-tuned NMT model frequently suffers from catastrophic failures, which the RAG-enhanced LLM effectively repairs. As illustrated in Table~\ref{tab:qualitative_examples} (see Appendix \ref{sec:appendix_qualitative}), we observed three distinct failure modes:

\paragraph{The "Safety Net" Effect}
The most prevalent NMT error is the \textbf{repetition loop}, where the model gets stuck generating a single token sequence (e.g., \textit{kahib'i-kalèbho} in GEN 10:17). The LLM demonstrates a language-agnostic ability to identify these non-sensical patterns, disregard the NMT draft, and re-translate based on the source and retrieved context.

\paragraph{Hallucination Correction}
The NMT model occasionally generates fluent but factually incorrect text. In GEN 11:5, the NMT hallucinates "King David" (\textit{dhèu aae Daud})—a figure frequent in the training data but absent in the source. The Post-Editor correctly identifies this mismatch against the English source ("The Lord") and the retrieved lexicon, correcting it to \textit{Lamatua}.

\paragraph{Syntactic Reconstruction}
Finally, the LLM successfully reconstructs complex genealogical idioms that confuse the NMT. In GEN 11:17, the NMT fails to render the phrase "became the father of," likely due to the structural divergence between the literal grammar book data and the idiomatic Bible. The Post-Editor, aided by retrieved examples, correctly utilizes the Dhao idiom \textit{matana} (`became the father of'), demonstrating the value of the word-level retrieval strategy in capturing high-frequency idiomatic patterns.

\subsection{Recommendations}
\label{sec:recommendations}
Based on our findings, we offer three key recommendations for practitioners working on unseen, low-resource languages:

\paragraph{Start with Context Volume, then look into Context Efficiency}
We recommend a two-step approach to RAG for low-resource MT: first, maximize the number of retrieved examples ($k$), as context saturation provides the most significant boost to translation quality regardless of the retrieval method. Second, tune for computational efficiency. Since our experiments show that distinct retrieval strategies converge to a similar performance band, practitioners can select the algorithm that best fits their latency constraints, switching from computationally expensive brute-force methods to faster alternatives, such as semantic search using sentence embedding models (e.g., BGE) or inverted indices (BM25), without sacrificing translation accuracy.

\paragraph{Consider Leaving Lexicographic Data for ICL, not Fine-Tuning}
Our ``NMT + Grammar'' baseline demonstrated that simply adding grammar book data at the fine-tuning stage can be detrimental. Performance actually degraded from 27.11 to 26.62 chrF++, likely because the rigid, pedagogical style of grammar book examples conflicts with the literary flow of the target domain. We show these resources can be best preserved as external knowledge bases for RAG, allowing the model to query specific terms dynamically without polluting the model's internal stylistic representations.

\paragraph{Keep an Out-of-Domain Test Set to Measure Robustness}
Standard practices in low-resource NMT often involve randomly splitting available corpora (e.g., the Bible) into training and test sets \citep{vazquez-etal-2021-helsinki, marashian-etal-2025-priest}. While many papers assume that the Bible belongs to a single, religious domain, our analysis shows a marked domain shift within this text, demonstrating that measuring out-of-domain performance is possible even when only the Bible is available as parallel corpus. Therefore, we recommend that low-resource researchers take this into consideration instead of using all verses of the Bible for both train and test, opting instead for document-level holdouts (e.g., distinct books or Testaments) to avoid inflated performance estimates \citep{khiu-etal-2024-predicting}. This aligns with recent findings from the WMT 2025 General Translation task \cite{kocmi-etal-2025-findings}, which argue that evaluating on "easy" in-domain data masks model brittleness and that robust assessment requires testing on challenging, document-level out-of-domain holdouts.

\section{Conclusion}
This work addresses domain shift in extremely low-resource settings. We demonstrate that while standard NMT suffers catastrophic degradation on unseen domains, our proposed hybrid NMT+LLM framework functions as a robust ``safety net,'' effectively recovering the quality lost to lexical and stylistic divergence.

Crucially, we find that context volume, rather than retrieval algorithm, is the primary driver of performance. We observe that distinct retrieval strategies (lexical vs. semantic) converge to comparable quality levels when normalized for volume. By validating these trade-offs on a language with no digital footprint, we provide a scalable blueprint for accelerating the translation of the Old Testament for thousands of low-resource languages worldwide.

\section{Limitations}
While this study provides a robust framework for tackling domain shift, it also highlights several limitations and clear avenues for future research:

\begin{enumerate}
    \item \textbf{Generalizability of Language and Domain:} The experiments were conducted on a single language pair (English-to-Dhao) and a single, specific domain shift (NT-to-OT). Future work is needed to test the generalizability of this framework. It would be valuable to validate whether the superiority of word-level retrieval and the "safety net" function of the LLM post-editor hold true for other low-resource language pairs and different types of domain shifts, such as translating from religious to secular text (e.g., news or health domains).

    \item \textbf{Optimizing Contextual Synergy:} As discussed in Section~\ref{sec:final_results}, our investigation into combining context types yielded mixed results. While combining the best sentence retrieval method with the full dictionary yielded the highest spBLEU score, it caused a slight decrease in the chrF++ score compared to using sentences alone. This suggests a lack of perfect synergy, likely because the large volume of combined data introduced noise. Future work should conduct finer-grained ablation studies to find the optimal balance, for instance, by combining parallel sentence retrieval with a \emph{retrieved subset} of the lexicon rather than the full dictionary, which may reduce noise and improve both metrics.
\end{enumerate}

\section*{Ethical Considerations}
\paragraph{Data Usage and Copyright.} This work utilizes data from the Dhao Alkitab 
(copyright \copyright 2012 Unit Bahasa dan Budaya) and \textit{A Grammar of Dhao} 
(Balukh, 2020). The Bible translation is licensed under \textbf{Creative Commons 
Attribution-NoDerivatives 4.0 International (CC BY-ND 4.0)}, which permits 
redistribution for research purposes provided the text remains unaltered. The 
grammar book is an \textbf{Open Access} publication (LOT Dissertation Series 570). 
Our use of these materials for non-commercial linguistic analysis and machine 
translation evaluation is consistent with these licenses and established 
\textbf{Fair Use} protocols for academic research. We provide full attribution 
to the original authors and rights holders in our citations.

\paragraph{Impact on Low-Resource Communities.} Our primary goal is to develop technologies that support the preservation and revitalization of very low-resource languages. We recognize that AI development for indigenous languages carries the risk of extractive research practices. To mitigate this, we focus on methods that can be deployed with minimal data and computational resources, making them accessible to local stakeholders. We hope this work serves as a foundation for future community-driven language tools.

\paragraph{Risks of Generative Models.} Neural Machine Translation and LLMs are prone to hallucinations, which poses a specific risk when handling sensitive or religious texts where accuracy is paramount. Our proposed \textbf{hybrid automated framework} (using LLMs as a post-editing safety net) is explicitly designed to identify and correct such anomalies. However, we emphasize that automated translations should always be reviewed by native speakers and community leaders before being treated as authoritative.

\section*{Acknowledgements}

This research was supported by the Commonwealth through an Australian Government Research Training Program Scholarship (\url{https://doi.org/10.82133/C42F-K220}). This research was undertaken using the LIEF HPC-GPGPU Facility hosted at the University of Melbourne. This Facility was established with the assistance of LIEF Grant LE170100200. Lau was supported by the Australian Research Council under Grant LP210200917. The authors also wish to thank Jermy Balukh for the use of his work, A Grammar of Dhao, which served as a foundational resource for our supplementary data extraction.

\bibliography{custom}

\appendix

\newpage

% ==========================================
% SECTION A: DOMAIN SHIFT ANALYSIS
% ==========================================
\section{Domain Shift Analysis}
\label{sec:domain_shift_analysis}

To quantify the lexical divergence between the New Testament (NT) and Old Testament (OT), we analyzed the frequency of domain-specific terms and the Out-of-Vocabulary (OOV) rates.

\vspace{1em}
\noindent
\begin{minipage}{\linewidth}
    \centering
    % Ensure filename matches your upload
    \includegraphics[width=\linewidth]{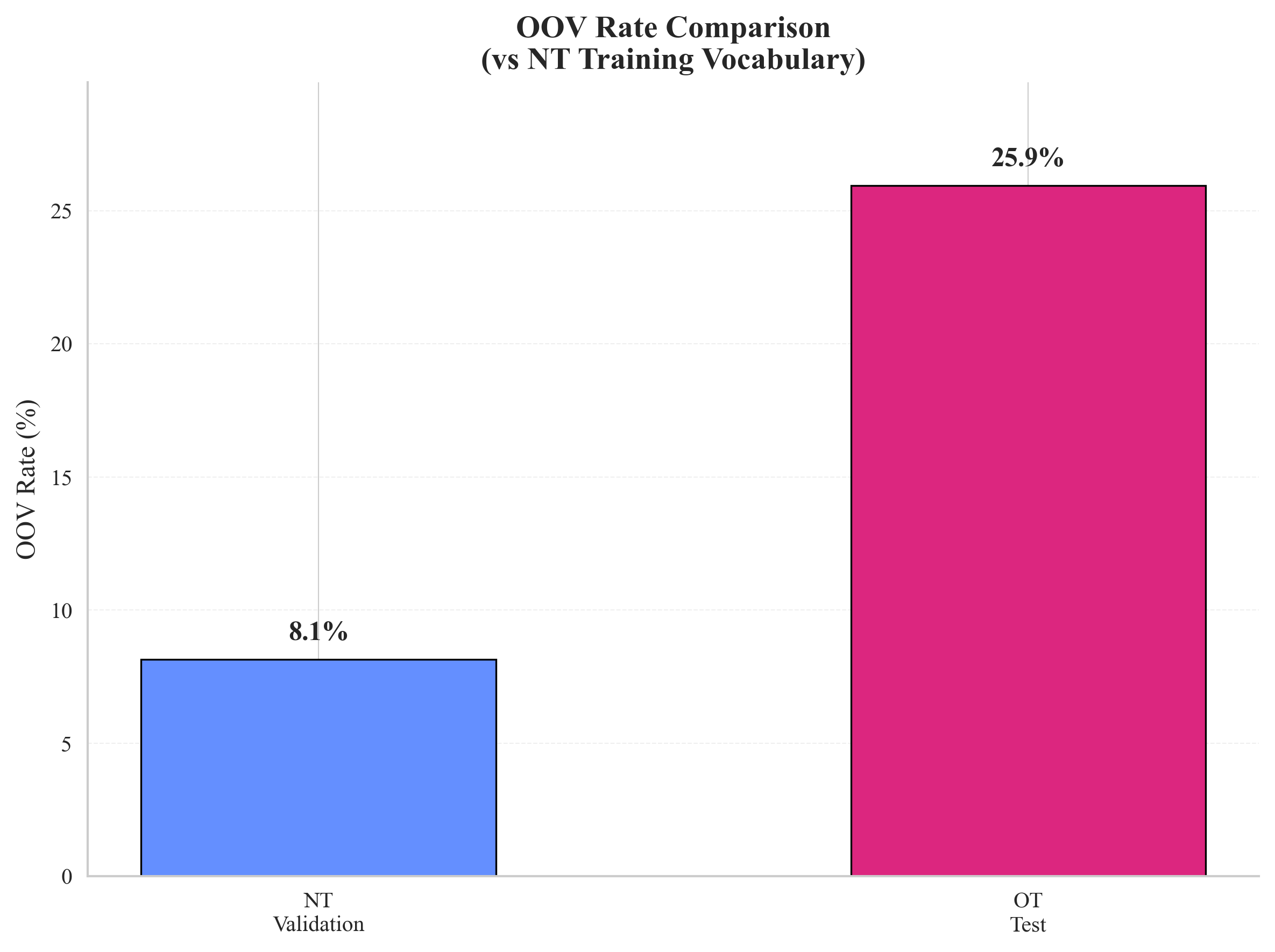}
    \captionof{figure}{The Out-of-Vocabulary (OOV) rate of the in-domain NT Validation set (8.1\%) versus the out-of-domain OT Test set (25.9\%). All rates are calculated relative to the NT training vocabulary.}
    \label{fig:oov_stats}
\end{minipage}
\vspace{1em}

\noindent
\begin{minipage}{\linewidth}
    \centering    
    % Ensure filename matches your upload
    \includegraphics[width=\linewidth]{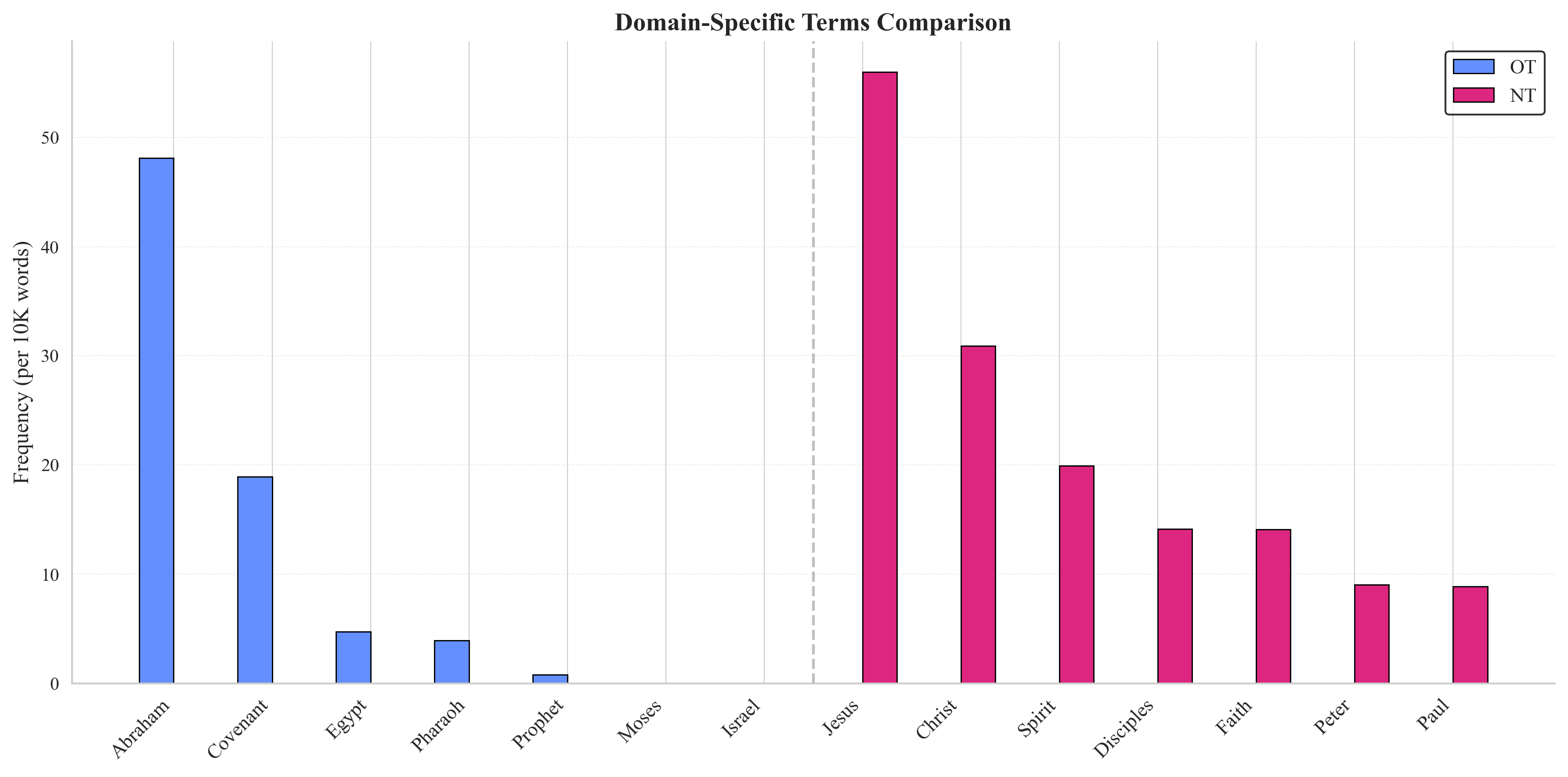}
    \captionof{figure}{Domain-specific term frequencies (normalized per 10k words) comparing the Old Testament (OT) and New Testament (NT) corpora. Note the prevalence of historical terms in the OT versus theological terms in the NT.}
    \label{fig:term_freq}
\end{minipage}

\vspace{1em}

\section{Data Construction Details}
\label{sec:data_construction}

We illustrate the complete data processing pipeline used to curate the experimental datasets from the unstructured grammar book and the raw biblical text.

\begin{figure*}[t]
    \centering
    \includegraphics[width=1.0\textwidth]{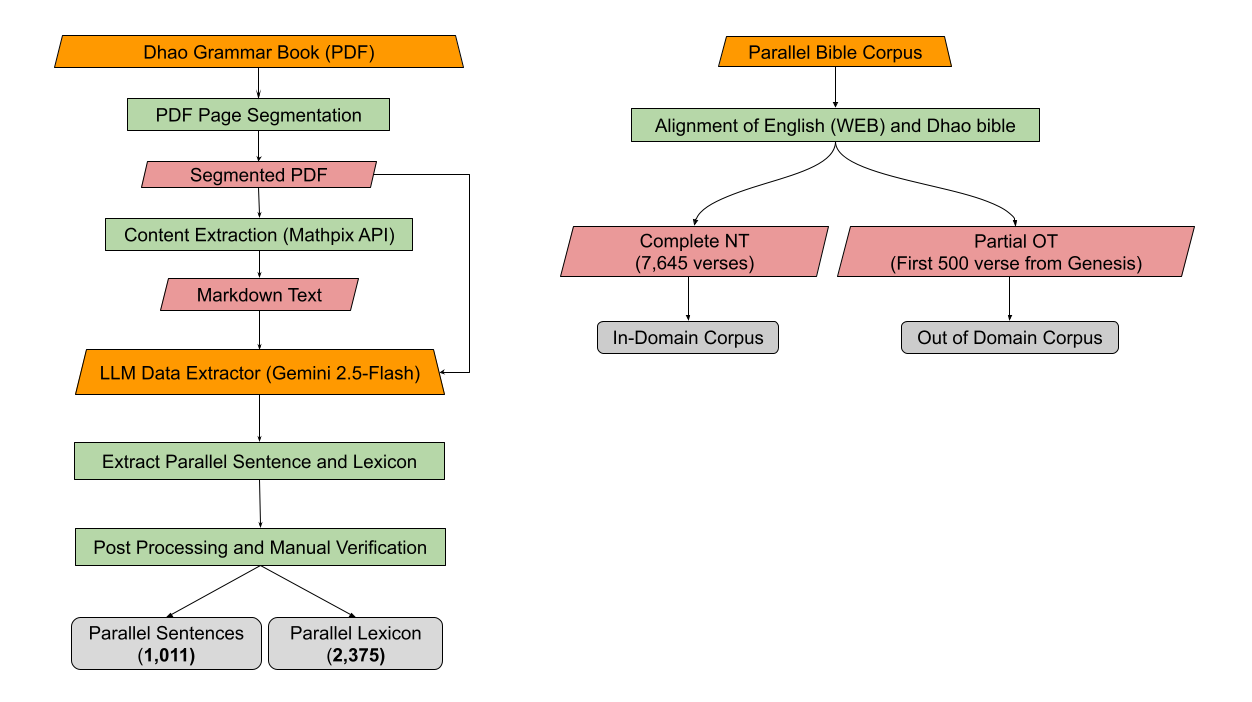}
    \caption{\textbf{Data Construction Pipeline.} The workflow processes two primary sources to create the experimental datasets. \textbf{Left:} We extract supplementary parallel sentences and lexicon entries from the unstructured Dhao Grammar Book PDF using a multi-stage pipeline involving OCR and LLM-based extraction. \textbf{Right:} The Parallel Bible Corpus is aligned and partitioned to simulate domain shift, using the New Testament (NT) as the in-domain training corpus and the Old Testament (OT) as the out-of-domain test set. \textbf{Legend:} Orange: Core objects (data and model) $|$ Green: Processing steps $|$ Red: Intermediate data $|$ Gray: Final output datasets.}
    \label{fig:data_pipeline}
\end{figure*}

\FloatBarrier
% ==========================================
% SECTION B: PROMPT TEMPLATES
% ==========================================
\section{Prompt Templates}
\label{sec:prompts}

To ensure reproducibility, we provide the exact prompt structures used for the Large Language Model (Gemini 2.5 Flash). We utilized two distinct prompt strategies: one for direct translation (baseline) and one for the post-editing task.

\subsection{System Instructions}

The following system prompts establish the persona and constraints for the LLM.

% DIRECT TRANSLATION PROMPT
\begin{tcolorbox}[breakable, colback=gray!5!white, colframe=gray!75!black, title=\textbf{Direct Translation Prompt}]\small
\textbf{System Message:} \\
Dhao is a member of the Sumba-Flores branch of the Malayo-Polynesian language family. It is spoken in Ndao Island in the Lesser Sunda Islands in Indonesia by about 5,000 people. It is classified as a member of the Sumba branch of Malayo-Polynesian languages, but may be a Papuan language. It is also known as Ndao, Ndaonese or Ndaundau.

You are an expert Bible translator in Dhao language. Your job is to translate bible verses from English to Dhao language, providing accurate and faithful translations that maintain the meaning and context of the source text. When provided with glossary entries or example translations, use them as reference to help ensure correct translation. You must respond ONLY with your translation in Dhao - no explanations, no reasoning, no additional text.

\rule{\textwidth}{0.4pt} % Separator

\textbf{User Template:} \\
Source text (English): \{src\_text\}

Translate the above text from English to Dhao:
\end{tcolorbox}

\vspace{1em}

% POST-EDITING PROMPT
\begin{tcolorbox}[breakable, colback=gray!5!white, colframe=gray!75!black, title=\textbf{Post-Editing Prompt}]\small
\textbf{System Message:} \\
Dhao is a member of the Sumba-Flores branch of the Malayo-Polynesian language family. It is spoken in Ndao Island in the Lesser Sunda Islands in Indonesia by about 5,000 people. It is classified as a member of the Sumba branch of Malayo-Polynesian languages, but may be a Papuan language. It is also known as Ndao, Ndaonese or Ndaundau.

You are an expert Bible translator in Dhao language. Your job is to correct and verify machine generated bible verses in Dhao language which is translated from the English language. Only make changes when necessary, ensuring that the post-edited dhao verse is aligned with the source English verse. When provided with glossary entries or example translations, use them as reference to help ensure correct translation. You must respond ONLY with the corrected translation text - no explanations, no reasoning, no additional text.

\rule{\textwidth}{0.4pt}

\textbf{User Template:} \\
Source text (English): \{src\_text\}

Machine translation (Dhao): \{pred\_text\}

Correct the machine translation if necessary:
\end{tcolorbox}

\subsection{Dynamic Context Injection}
Depending on the retrieval strategy, the following blocks are dynamically appended to the User Template.

% FEW-SHOT EXAMPLES BOX
\begin{tcolorbox}[breakable, colback=blue!5!white, colframe=blue!75!black, title=\textbf{Dynamic Component: Parallel Sentence Examples}]\small
\textit{(Appended when $k > 0$ parallel sentences are retrieved)}

To help with the translation, here are some example parallel sentences between Dhao and English:

Dhao: [target\_example\_1] \\
English translation: [source\_example\_1]

Dhao: [target\_example\_2] \\
English translation: [source\_example\_2]
...
\end{tcolorbox}

% GLOSSARY BOX
\begin{tcolorbox}[breakable, colback=green!5!white, colframe=green!75!black, title=\textbf{Dynamic Component: Glossary Entries}]\small
\textit{(Appended when Lexicon Retrieval is enabled)}

To help with the translation, here is a word list between English and Dhao in the format: English word (pos tag) -> Dhao word:
\begin{itemize}
    \item source\_word\_1 (noun) -> target\_word\_1
    \item source\_word\_2 (verb) -> target\_word\_2
    \item source\_word\_3 -> target\_word\_3
\end{itemize}
\end{tcolorbox}
\vspace{1em}

% ==========================================
% SECTION D: HYPERPARAMETERS CONFIGURATION
% ==========================================
\newpage
\section{Hyperparameters and Training Details}
\label{sec:appendix_hyperparams}
All NMT models were fine-tuned using the Hugging Face \texttt{transformers} library on a single NVIDIA A100 GPU. We utilized the \texttt{facebook/nllb-200-distilled-600M} pretrained checkpoint. Table \ref{tab:hyperparameters} summarizes the hyperparameters used for both the baseline (NMT-Only) and the augmented (NMT + Grammar) configurations.

Note that the \textit{NMT + Grammar} model was trained for more steps (7,000 vs 5,000) to account for the additional training data provided by the grammar book CSV.

\vspace{2em}

\begin{minipage}{\linewidth}
    \centering
    \small
    \resizebox{\columnwidth}{!}{%
    \begin{tabular}{l c c}
        \toprule
        \textbf{Hyperparameter} & \textbf{NMT-Only} & \textbf{NMT + Grammar} \\
        \midrule
        Base Model & \multicolumn{2}{c}{NLLB-200-distilled-600M} \\
        Precision & \multicolumn{2}{c}{bfloat16} \\
        Attention Impl. & \multicolumn{2}{c}{SDPA (Scaled Dot Product Attention)} \\
        \midrule
        Learning Rate & \multicolumn{2}{c}{2e-4} \\
        Label Smoothing & \multicolumn{2}{c}{0.2} \\
        Warmup Steps & \multicolumn{2}{c}{1,000} \\
        Early Stopping Patience & \multicolumn{2}{c}{4} \\
        \midrule
        Batch Size (per device) & \multicolumn{2}{c}{16} \\
        Grad. Accumulation Steps & \multicolumn{2}{c}{4} \\
        Effective Batch Size & \multicolumn{2}{c}{64} \\
        \midrule
        Max Sequence Length & \multicolumn{2}{c}{400} \\
        Beam Size (Eval) & \multicolumn{2}{c}{2} \\
        \midrule
        \textbf{Max Training Steps} & \textbf{5,000} & \textbf{7,000} \\
        \bottomrule
    \end{tabular}%
    }
    \captionof{table}{Fine-tuning hyperparameters for the NMT baselines.}
    \label{tab:hyperparameters}
\end{minipage}

\vspace{2em}
% ==========================================
% SECTION E: DETAILED RETRIEVAL RESULTS
% ==========================================
\newpage
\section{Detailed Retrieval Results}
\label{sec:appendix_results}

We provide the complete numerical results for the retrieval ablation studies discussed in Section \ref{sec:retrieval_analysis}. Table \ref{tab:app_sentence_results} details the performance of sentence-level strategies (BM25, BGE, ChrF-RAG) across varying context sizes ($K$). Table \ref{tab:app_word_results} details the word-level fuzzy matching strategy across varying retrieval densities ($N$). Finally, Table \ref{tab:app_lexicon_results} presents the results for Lexicon Retrieval, comparing dynamic retrieval against the static full-dictionary baseline.

% TABLE 1: SENTENCE LEVEL
\begin{table}[htbp]
    \centering
    \small
    \resizebox{\linewidth}{!}{%
    \begin{tabular}{llcc}
        \toprule
        \textbf{Method} & \textbf{K} & \textbf{spBLEU} & \textbf{chrF++} \\
        \midrule
        \multicolumn{4}{l}{\textit{Baseline}} \\
        NMT (NLLB) & - & 7.66 & 27.11 \\
        \midrule
        \multicolumn{4}{l}{\textit{BGE (Semantic)}} \\
        BGE & 5 & 15.23 & 31.98 \\
        BGE & 10 & 15.71 & 32.45 \\
        BGE & 20 & 16.97 & 33.50 \\
        BGE & 40 & 17.34 & 33.88 \\
        BGE & 60 & 18.34 & 34.72 \\
        BGE & 80 & 18.32 & 34.83 \\
        BGE & 100 & 18.47 & 34.87 \\
        \midrule
        \multicolumn{4}{l}{\textit{BM25 (Lexical)}} \\
        BM25 & 5 & 14.76 & 31.65 \\
        BM25 & 10 & 15.50 & 32.42 \\
        BM25 & 20 & 16.53 & 33.31 \\
        BM25 & 40 & 17.44 & 34.01 \\
        BM25 & 60 & 17.34 & 33.97 \\
        BM25 & 80 & 17.51 & 34.28 \\
        BM25 & 100 & 17.25 & 34.19 \\
        \midrule
        \multicolumn{4}{l}{\textit{ChrF-RAG (Diversity)}} \\
        ChrF & 5 & 15.33 & 32.09 \\
        ChrF & 10 & 16.32 & 32.94 \\
        ChrF & 20 & 17.01 & 33.74 \\
        ChrF & 40 & 18.05 & 34.61 \\
        \textbf{ChrF} & \textbf{60} & \textbf{18.44} & \textbf{34.98} \\
        ChrF & 80 & 18.12 & 34.76 \\
        ChrF & 100 & 18.27 & 34.93 \\
        \bottomrule
    \end{tabular}%
    }
    \caption{Detailed ablation results for \textbf{Sentence-Level} retrieval methods. Note that performance gains tend to plateau around $K=60-80$ for most methods.}
    \label{tab:app_sentence_results}
\end{table}

% TABLE 2: WORD LEVEL
% Changed [H] to [htbp]
\begin{table}[htbp]
    \centering
    \small
    \resizebox{\linewidth}{!}{%
    \begin{tabular}{lccc}
        \toprule
        \textbf{N (per word)} & \textbf{Eff. K} & \textbf{spBLEU} & \textbf{chrF++} \\
        \midrule
        \multicolumn{4}{l}{\textit{Word-Level Fuzzy Matching}} \\
        1 & $\approx$14 & 15.96 & 32.87 \\
        2 & $\approx$28 & 16.94 & 33.64 \\
        3 & $\approx$42 & 17.72 & 34.42 \\
        5 & $\approx$70 & 18.72 & 35.00 \\
        \textbf{137} & \textbf{$\approx$137} & \textbf{18.93} & \textbf{35.28} \\
        15 & $\approx$205 & 18.82 & 35.16 \\
        20 & $\approx$272 & 16.90 & 34.22 \\
        \bottomrule
    \end{tabular}%
    }
    \caption{Detailed ablation for the \textbf{Word-Level} strategy. "Eff. K" denotes the approximate effective number of sentences retrieved. Performance peaks at $N=10$ before degrading due to context noise.}
    \label{tab:app_word_results}
\end{table}

% TABLE 3: LEXICON RETRIEVAL
% Changed [H] to [htbp]
\begin{table}[htbp]
    \centering
    \small
    \resizebox{\linewidth}{!}{%
    \begin{tabular}{lccc}
        \toprule
        \textbf{N (per word)} & \textbf{Eff. K} & \textbf{spBLEU} & \textbf{chrF++} \\
        \midrule
        \multicolumn{4}{l}{\textit{Word-Level Fuzzy Matching}} \\
        3 & $\approx$77 & 13.84 & 29.56 \\
        5 & $\approx$128 & 14.29 & 29.76 \\
        10 & $\approx$256 & 14.18 & 29.66 \\
        15 & $\approx$384 & 14.53 & 29.84 \\
        20 & $\approx$512 & 14.85 & 29.84 \\
        25 & $\approx$640 & 15.19 & 30.25 \\
        30 & $\approx$768 & 15.19 & 30.42 \\
        50 & $\approx$1280 & 15.29 & 30.58 \\
        70 & $\approx$1791 & 15.76 & 30.61 \\
        100 & $\approx$2559 & 16.22 & 30.88 \\
        \midrule
        \multicolumn{4}{l}{\textit{Static Context}} \\
        \textbf{Full Dictionary} & \textbf{2377} & \textbf{16.27} & \textbf{31.32} \\
        \bottomrule
    \end{tabular}%
    }
    \caption{Detailed ablation for Lexicon Retrieval. Unlike sentence retrieval, performance scales monotonically with volume, peaking when the Full Dictionary is provided.}
    \label{tab:app_lexicon_results}
\end{table}

\FloatBarrier

\section{Qualitative Analysis Examples}
\label{sec:appendix_qualitative}

\begin{minipage}{2\linewidth}
We provide concrete examples of the failure modes discussed in Section \ref{sec:qualitative_analysis}. Table \ref{tab:qualitative_examples} highlights three specific instances where the NMT baseline failed catastrophically on the Out-of-Domain test set, and how the RAG-enhanced Post-Editor recovered the correct translation.

\vspace{1em}
    \centering
    \small
    \resizebox{\textwidth}{!}{%
    \begin{tabular}{l p{5cm} p{5cm} p{5cm}}
        \toprule
        \textbf{Type} & \textbf{Source (English)} & \textbf{NMT Output (Draft)} & \textbf{LLM Post-Edit (Final)} \\
        \midrule
        \textbf{Repetition Loop} & \textit{"it rained on the earth forty days and forty nights"} (GEN 7:12) & \textit{Hèia bèli-camèd'a bèli-camèd'a...} (repeats indefinitely) & \textit{Èj'i bhori ètu rai èpa nguru lod'o mèu-mèda.} (Correct translation) \\
        \midrule
        \textbf{Hallucination} & \textit{"the lord came down to see the city..."} (GEN 11:5) & \textit{...dhèu aae \textbf{Daud} puru...} ("King \textbf{David} came down...") & \textit{\textbf{Lamatua} puru nèti dedha mai...} ("\textbf{The Lord} came down...") \\
        \midrule
        \textbf{Complex Syntax} & \textit{"...after he became the father of peleg"} (GEN 11:17) & \textit{Nèti èèna ka, Eber mamuri toke d'ai... Èle èèna ka, nèti èèna ka, nèti èèna ka...} (Degenerates into repetition loop) & \textit{...èli nèngu \textbf{matana} Peleg...} (Recovers genealogical idiom) \\
        \bottomrule
    \end{tabular}%
    }
    \captionof{table}{Examples of NMT failures corrected by the RAG-enhanced Post-Editor. The LLM acts as a safety net, fixing repetition loops, named entity hallucinations, and recovering complex idioms.}
    \label{tab:qualitative_examples}
\end{minipage}

\end{document}